\pdfoutput=1
\documentclass{article}
\usepackage{ijcai24}

\usepackage{times}
\usepackage{soul}
\usepackage{url}
\usepackage[hidelinks]{hyperref}
\usepackage[utf8]{inputenc}
\usepackage[small]{caption}
\usepackage{graphicx}
\usepackage{amsmath}
\usepackage{amsthm}
\usepackage{booktabs}
\usepackage{algorithm}
\usepackage{algorithmic}
\usepackage[switch]{lineno}

\usepackage{newfloat}
\usepackage{listings}
\usepackage{multirow}
\usepackage{amsmath}
\usepackage{amssymb}
\usepackage{booktabs} 
\usepackage{makecell}
\usepackage{bbding}
\usepackage{color}

\title{Invertible Residual Rescaling Models}

\author{
Jinmin Li$^1$
\and
Tao Dai$^{2,}$\thanks{Corresponding author: Tao Dai (daitao.edu@gmail.com)}
\and
Yaohua Zha$^{1,3}$
\and
Yilu Luo$^1$ 
\and
Longfei Lu$^1$ 
\and
Bin Chen$^4$ 
\and
Zhi Wang$^1$ 
\and
Shu-Tao Xia$^{1,3}$
\and
Jingyun Zhang$^5$
\affiliations
$^1$Tsinghua Shenzhen International Graduate School, Tsinghua University
\\
$^2$College of Computer Science and Software Engineering, Shenzhen University\\
$^3$Research Center of Artificial Intelligence, Peng Cheng Laboratory\\
$^4$Department of Computer Science and Technology, Harbin Institute of Technology, Shenzhen \\
$^5$WeChat Pay Lab33, Tencent\\
\emails
\{ljm22, chayh21, luo-yl22\}@mails.tsinghua.edu.cn,
\{daitao.edu, loneffy.lu, zhang304973926\}@gmail.com,
chenbin2021@hit.edu.cn,
\{wangzhi, xiast\}@sz.tsinghua.edu.cn
}

\begin{document}

\maketitle

\begin{abstract}
    Invertible Rescaling Networks (IRNs) and their variants have witnessed remarkable achievements in various image processing tasks like image rescaling. However, we observe that IRNs with deeper networks are difficult to train, thus hindering the representational ability of IRNs. To address this issue, we propose Invertible Residual Rescaling Models (IRRM) for image rescaling by learning a bijection between a high-resolution image and its low-resolution counterpart with a specific distribution. Specifically, we propose IRRM to build a deep network, which contains several Residual Downscaling Modules (RDMs) with long skip connections. Each RDM consists of several Invertible Residual Blocks (IRBs) with short connections. In this way, RDM allows rich low-frequency information to be bypassed by skip connections and forces models to focus on extracting high-frequency information from the image. Extensive experiments show that our IRRM performs significantly better than other state-of-the-art methods with much fewer parameters and complexity. Particularly, our IRRM has respectively PSNR gains of at least 0.3 dB over HCFlow and IRN in the $\times 4$ rescaling while only using 60\% parameters and 50\% FLOPs. The code will be available at https://github.com/THU-Kingmin/IRRM.
\end{abstract}

\section{Introduction}
    Image rescaling, which aims to reconstruct high-resolution (HR) images from their corresponding low-resolution (LR) versions by forward sampling, has been widely utilized in large-size data services like storage and transmission.  Recently, many flow-based generative methods~\cite{flow1,flow2,flow3,IRN,HCFlow} have been applied in image generation and achieved remarkable progress in image rescaling and image super-resolution (SR). 
    
    \begin{figure}[ht]
        \centering
        \includegraphics[width=\linewidth]{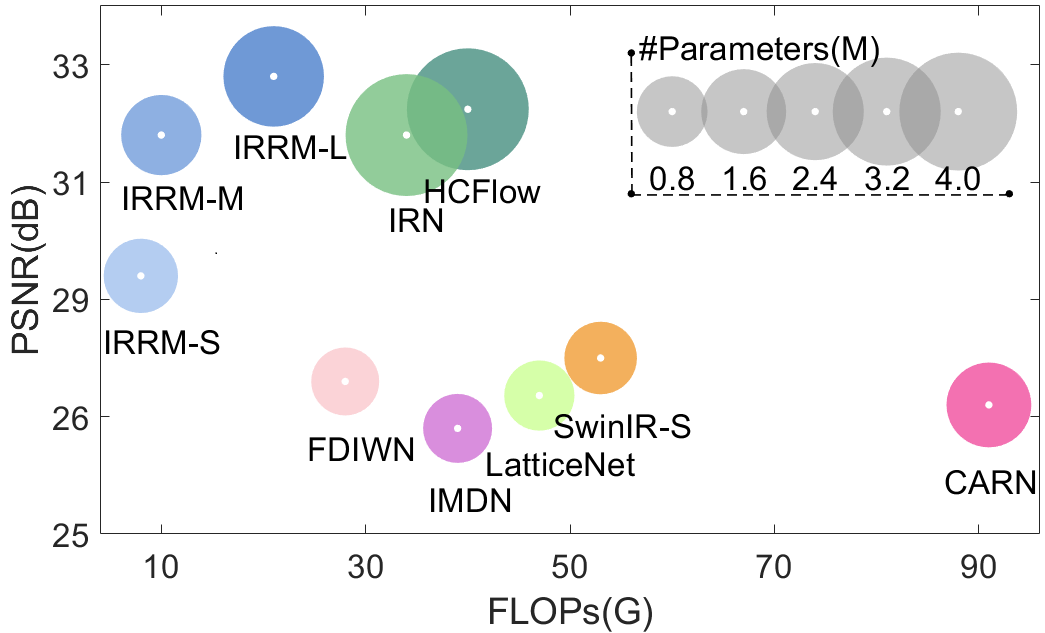}
        \caption{Comparison results of state-of-the-art methods (like IRN and HCFlow) and our IRRM on Urban100 dataset with ×4 rescaling. SR methods combined with bicubic downscaling are also reported. -S, -M, and -L represent the sizes of model parameters respectively, where -S represents small, -M represents medium, and -L represents large. IRRM achieves similar performance using a quarter of parameters and FLOPs of the IRN and HCFlow.}
        \label{figs:FLOPs_PSNR}
    \end{figure}
    
    Previous SR methods focus on non-adjustable downscaling kernel (e.g., Bicubic interpolation) rescaling, while omitting their compatibility with the downscaling operation. To remedy this issue, more recent works  design adjustable downscaling networks and  treat image rescaling as a unified task through an encoder-decoder framework~\cite{IR1,IR2}. 
    Specifically, they suggest using an optimal downscaling method as an encoder, trained together with an existing SR module. Despite the improvement of reconstruction quality, these methods still face the problem of the recovery of high-frequency information.
    
    To restore high-frequency details better, Invertible Rescaling Network (IRN)~\cite{IRN} has been developed and achieved impressive performance in image rescaling.  This type of method focuses on learning the downscaling and upscaling with an invertible neural network, which captures the lost information in a specific distribution and embeds it in the parameters of the model.
    More recently,  HCFlow~\cite{HCFlow} proposes an invertible conditional flow-based model to model the HR-LR relationship, where the high-frequency component is hierarchically conditional on the low-frequency component of the image. Despite the success of IRN in image rescaling, such IRNs are difficult to train, especially with deep layers.

    To solve this problem, in this paper, we propose a novel Invertible Residual Rescaling Model (IRRM) to facilitate training while enhancing the representational ability of models. To ease the training, we propose Residual Downscaling Module (RDM) with long skip connections, which serve as the basic block of our IRRM and long skip connection allows rich low-frequency information to be bypassed. In each RDM module, we stack several invertible residual blocks (IRB) to enhance non-linear representation and reduce model degradation with short skip connections. With long and short skip connections, abundant information can be bypassed and thus ease the flow of information.
    As shown in Fig.~\ref{figs:FLOPs_PSNR}, our IRRM obtains better results than other state-of-the-art methods with much fewer parameters and complexity.
    
    In summary, the main contributions are as follows:
    \begin{itemize}
        \item We propose a novel Invertible Residual Rescaling Model (IRRM) to build deep networks for highly accurate image rescaling. Our IRRM obtains much better performance than previous methods with much fewer parameters and complexity.
        \item We propose Residual Downscaling Module (RDM) with long skip connections, equivalent to the second-order wavelet transform, to allow the model to focus on learning the texture information of the image while easing the flow of information. 
        \item Our proposed IRRM introduces the Invertible Residual Block (IRB), which incorporates short skip connections to enhance the model's nonlinear representational ability. This addition significantly improves the extensibility of the model.
    \end{itemize}
    
\section{Related Work}
    \subsection{Image Super-Resolution}
        Image super-resolution is a different task from rescaling, aiming to restore HR image given the LR image. However, SR combined with downsampling can be used for image rescaling task. Recently, SR networks have achieved impressive performance based on deep learning~\cite{CFGN,FSR,cui2023image,zhang2023lightweight,HPNet,mambair}. SRCNN~\cite{SRCNN} is the first work that applies convolutional neural networks  for image SR. Later, many methods~\cite{VDSR,SRResNet} stack many convolutional layers with residual connections to facilitate the network training. Recently, other advanced methods like RCAN~\cite{RCAN}, SAN \cite{SAN} and SwinIR \cite{SwinIR} build very deep networks to enhance the feature expression ability and have obtained remarkable performance with substantial computational cost.
        On the other hand, light-weight SR works have been proposed to relieve the problem of computational cost~\cite{IMDN,LatticeNet}. For example, the lightweight frequency-aware network (FADN)~\cite{FADN2021ICCV} is developed to reconstruct different frequency signals with various operations. Instead of designing a lightweight SR network, ClassSR~\cite{ClassSR} develops a general framework to accelerate SR networks. However, they seem to have reached the limits of performance for image generation tasks.
        
    \subsection{Image Rescaling}
        The goal of image rescaling is to downscale a high-resolution (HR) image into a visually satisfying low-resolution (LR) image and then restore the HR image from the LR image. Typically, image rescaling concurrently models the enlargement and reduction processes within an encoder-decoder structure to fine-tune the reduction model for subsequent enlargement tasks~\cite{IR1,IR2,baflow}.
        Recently, IRN~\cite{IRN} has been introduced to model the rescaling process using a bijective invertible neural network. During the training phase, IRN effectively captures the high-frequency and low-frequency components. During the testing phase, the HR image can be reconstructed by the generated LR image and a randomly sampled Gaussian distribution. Besides, HCFlow~\cite{HCFlow} assumes that the high-frequency component is dependent on the LR image and employs a hierarchical conditional framework to model the LR image from conditional distribution of the high-frequency component.
        Nevertheless, these models cannot extend to larger models because of the excessive focus on low frequency information and the lack of residual connections. We address these issues by introducing residual connections and residual enhanced block equivalent to the second-order wavelet transform.

        \begin{figure}
            \centering
            \includegraphics[width=0.8\linewidth]{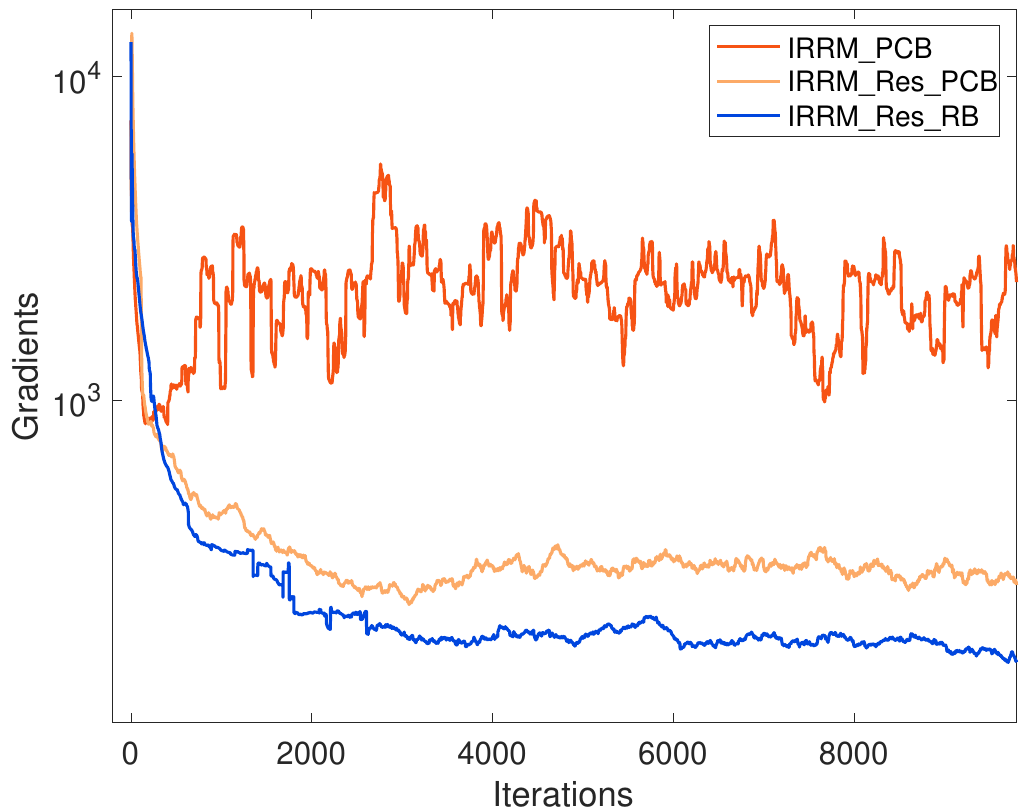}
            \caption{Illustration of training gradients for different models. IRRM with residual connections and enhanced residual block ($\text{IRRM\_Res\_RB}$) gain a more stable gradient than IRRM w/o residual connections ($\text{IRRM\_PCB}$), leading to faster convergence and better performance.}
            \label{figs:gradient_compa}
        \end{figure}
        
    \begin{figure*}
        \centering
        \includegraphics[width=0.9\linewidth]{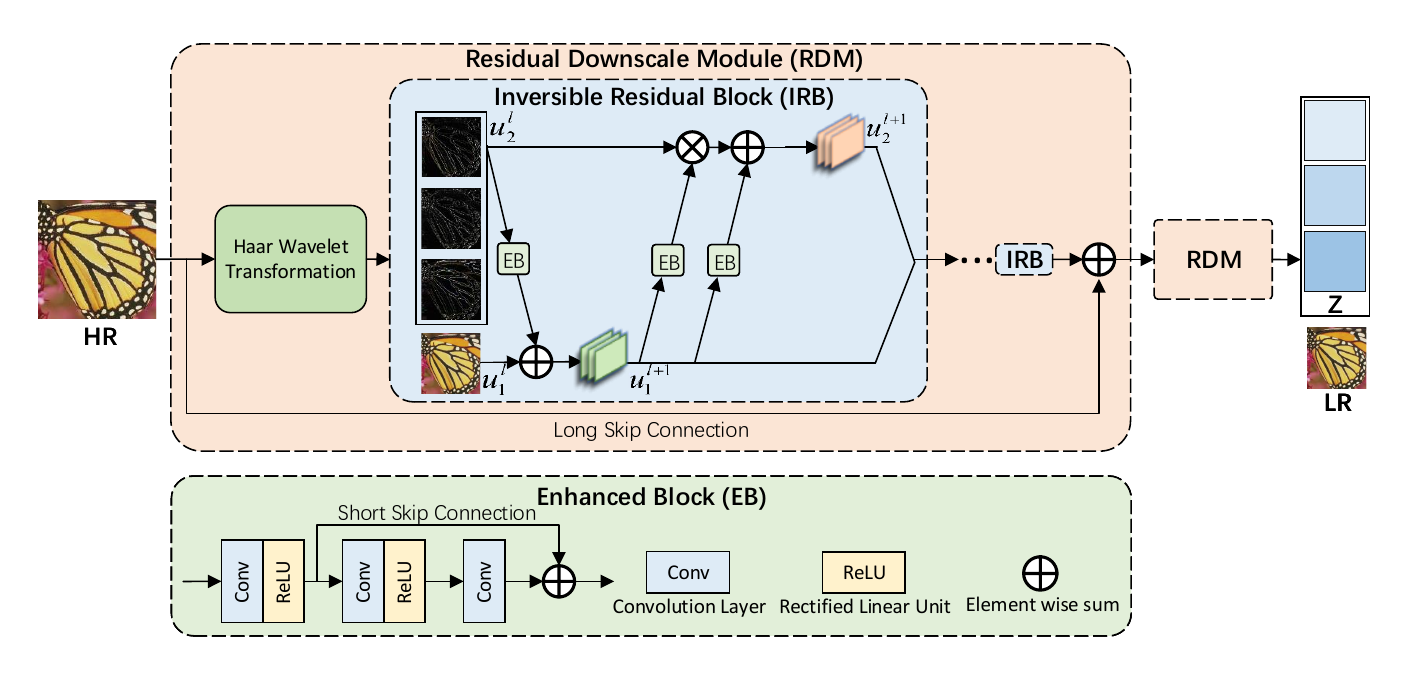}
        \caption{The overall framework of Invertible Residual Rescaling Models (IRRM). IRRM is composed of Residual Downscaling Modules (RDMs), in which Invertible Residual Blocks (IRBs) are stacked after a wavelet transformation. Each IRB contains three Enhanced Blocks (EBs) to enhance the nonlinear representation and mitigate vanishing the gradient problem.}
        \label{figs:IRRM}
    \end{figure*}

\section{Methodology}
    \subsection{Model Specification}
        Image rescaling involves reconstructing a high-resolution (HR) image $x$ from a low-resolution (LR) image $y$ that is obtained by downscaling $x$. To achieve this, we employ an invertible neural network that models the LR image $y$ and ensures that the distribution of the lost information conforms to a predetermined distribution $z$ (denoted as $[y, z] = f_\theta(x)$, where $z$ is sampled from $p(z)$ and $x$ follows the distribution $p(x)$). Conversely, we can reconstruct $x$ through the inverse process using $[y, z]$, denoted as $x = f_\theta^{-1}(y, z)$. It is important to note that the lost information $z$ corresponds to the high-frequency details of the image, as per the Nyquist-Shannon sampling theorem~\cite{NyShan}. Instead of explicitly preserving the high-frequency information, we can model $z$ as a Gaussian distribution, thereby obtaining it through sampling from a Gaussian distribution.
        
    \subsection{Invertible Residual Rescaling Models}
        As shown in Fig.~\ref{figs:IRRM}, our proposed IRRM consists of stacked Residual Downscaling Module (RDM), where RDM further contains one Haar Wavelet Transformation block and several Invertible Residual Blocks (IRBs).
        
        \noindent{\bf Residual Downscaling Module.} RDM utilizes Haar transformation~\cite{FDWT} to split the input into high and low frequency information. Then IRB learn to model these frequency information into the specified distribution. Specifically, given an HR image $x$ with shape $(H, W, C)$, we first obtain the residual representation of $x$:
        \begin{equation}
            x_{res} = x - upsample(y)
        \end{equation}
        where $y$ is the LR image and $y = downsample(x)$. Then wavelet transformation decompose $x_{res}$ into global frequency features $[u_1^k, u_2^k]$:
        \begin{equation}
            \begin{aligned}
                \relax[u_1^k, u_2^k] &= \text{WaveletTransform}(x_{res}^k)\\
            \end{aligned}
        \end{equation}
        where $k$ denotes $k^{th}$ layer of the model. $u_1$ and $u_2$ correspond to high-frequency and low-frequency information, respectively.
        $x_{res}$ facilitates the model to focus on learning high frequency representations, aiming to recover textures of the image, which is equivalent to the second-order wavelet transform (IRN~\cite{IRN} is the first-order wavelet transform). Meanwhile, residual connections make the training process more stable in favour of convergence, as shown in Fig.~\ref{figs:gradient_compa}. IRB extracts image features from $u_1^k$ and $u_2^k$ to $u_1^{k+1}$ and $u_2^{k+1}$. After n IRBs' transformation, we obtain the downscaled image $y$ with shape $(\frac{1}{2}H, \frac{1}{2}W, C)$ and Gaussian distribution $z$ with shape $(\frac{1}{2}H, \frac{1}{2}W, 3C)$.

        \begin{figure}[t]
            \centering
            \includegraphics[width=0.9\linewidth]{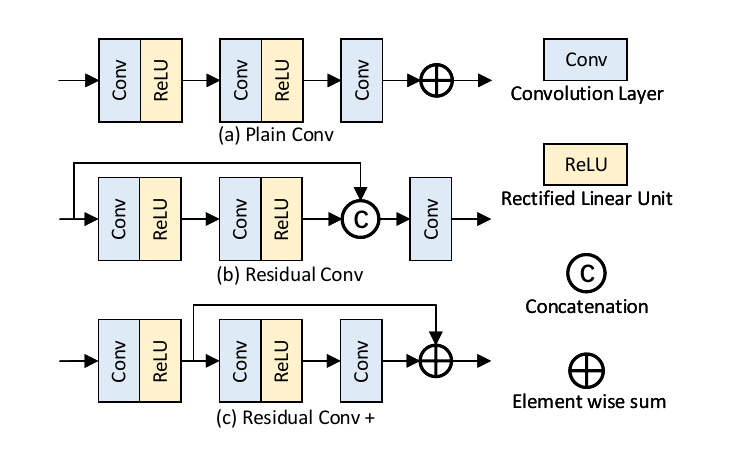}
            \caption{Illustration of Enhanced Blocks (EBs). Three non-linear convolutional blocks in the Invertible Residual Block are compared.}
            \label{figs:EB}
        \end{figure}

   \begin{table*}[ht]
    \small
    \centering
    \begin{tabular}{c|c|c|c|c|c|c|c}
    \hline \multirow{2}{*}{ Scale } & \multirow{2}{*}{Models} & \multirow{2}{*}{Paras} & Set5 & Set10  & B100 & Urban100 & DIV2K \\
    \cline{4-8}& & & PSNR / SSIM & PSNR / SSIM& PSNR / SSIM& PSNR / SSIM& PSNR / SSIM \\
    \hline \multirow{14}{*}{$\times 4$} 
    & Bicubic \& Bicubic &- & 28.42 / 0.8104 & 26.00 / 0.7027 & 25.96 / 0.6675 & 23.14 / 0.6577 & 26.66 / 0.8521 \\
     & Bicubic \&SRCNN & $57.3 \mathrm{~K}$ & 30.48 / 0.8628 & 27.50 / 0.7513 & 26.90 / 0.7101 & 24.52 / 0.7221 & - / - \\
     & Bicubic \& CARN & $1.59 \mathrm{M}$ & 32.13 / 0.8937 & 28.60 / 0.7806 & 27.58 / 0.7349 & 26.07 / 0.7837 & - / - \\
     & Bicubic \& EDSR & $43.1 \mathrm{M}$ & 32.46 / 0.8968 & 28.80 / 0.7760 & 27.71 / 0.7420 & 26.64 / 0.8033 & 29.38 / 0.9032 \\
     & Bicubic \& RCAN & $15.6 \mathrm{M}$ & 32.63 / 0.9002 & 28.87 / 0.7889 & 27.77 / 0.7436 & 26.82 / 0.8087 & 30.77 / 0.8460 \\
     & Bicubic \& SAN & $15.7 \mathrm{M}$ & 32.64 / 0.9003 & 28.92 / 0.7888 & 27.78 / 0.7436 & 26.79 / 0.8068 & - / - \\
     & Bicubic \& SwinIR & $11.9 \mathrm{M}$ & 32.92 / 0.9044 & 29.09 / 0.7950 & 27.92 / 0.7489 & 27.45 / 0.8254 & - / - \\
     & Bicubic \& HAT & $20.8 \mathrm{M}$ & 33.30 / 0.9083 & 29.47 / 0.8015 & 28.09 / 0.7551 & 28.60 / 0.8498 & - / - \\
    
    \cline{2-8} & TAD \& TAU & - & 31.81 /  - & 28.63 /  - & 28.51 /  - & 26.63 /  - & 31.16 /  - \\
     & CAR \& EDSR & $51.1 \mathrm{M}$ & 33.88 / 0.9174 & 30.31 / 0.8382 & 29.15 / 0.8001 & 29.28 / 0.8711 & 32.82 / 0.8837 \\
     & IRN & $4.35 \mathrm{M}$ & 36.19 / 0.9451 & 32.67 / 0.9015 & 31.64 / 0.8826 & 31.41 / 0.9157 & 35.07 / 0.9318 \\
     & HCFlow & $4.35 \mathrm{M}$ & 36.29 / 0.9468 & \textcolor{blue}{33.02} / 0.9065 & 31.74 / 0.8864 & \textcolor{blue}{31.62 / 0.9206} & 35.23 / 0.9346 \\
     & \textbf{IRRM-S (ours)} & $991 \mathrm{~K}$ & 33.20 / 0.8927 & 29.75 / 0.8441 & 30.82 / 0.8547 & 30.21 / 0.8825 & 33.45 / 0.8901 \\
     & \textbf{IRRM-M (ours)} & $1.32 \mathrm{M}$ & \textcolor{blue}{36.30 / 0.9491} & 32.84 / \textcolor{blue}{0.9098} & \textcolor{blue}{31.93 / 0.8938} & 31.36 / 0.9193 & \textcolor{blue}{35.31 / 0.9373} \\
     & \textbf{IRRM-L (ours)} & $2.64 \mathrm{M}$ & \textcolor{red}{36.85 / 0.9541} & \textcolor{red}{33.37 / 0.9174} & \textcolor{red}{32.30 / 0.9016} & \textcolor{red}{31.94 / 0.9278} & \textcolor{red}{35.78 / 0.9429} \\
    
    \hline \multirow{13}{*}{$\times 2$} 
     & Bicubic \& Bicubic & - & 33.66 / 0.9299 & 30.24 / 0.8688 & 29.56 / 0.8431 & 26.88 / 0.8403 & 31.01 / 0.9393 \\
     & Bicubic \& SRCNN & $57.3 \mathrm{~K}$ & 36.66 / 0.9542 & 32.45 / 0.9067 & 31.36 / 0.8879 & 29.50 / 0.8946 & -  /  - \\
     & Bicubic \& CARN & $1.59 \mathrm{M}$ & 37.76 / 0.9590 & 33.52 / 0.9166 & 32.09 / 0.8978 & 31.92 / 0.9256 & -  /  - \\
     & Bicubic \& EDSR & $40.7 \mathrm{M}$ & 38.20 / 0.9606 & 34.02 / 0.9204 & 32.37 / 0.9018 & 33.10 / 0.9363 & 35.12 / 0.9699 \\
     & Bicubic \& RCAN & $15.4 \mathrm{M}$ & 38.27 / 0.9614 & 34.12 / 0.9216 & 32.41 / 0.9027 & 33.34 / 0.9384 & -  /  - \\
     & Bicubic \& SAN & $15.7 \mathrm{M}$ & 38.31 / 0.9620 & 34.07 / 0.9213 & 32.42 / 0.9208 & 33.10 / 0.9370 & -  /  - \\
     & Bicubic \& SwinIR & $11.8 \mathrm{M}$ & 38.42 / 0.9623 & 34.46 / 0.9250 & 32.53 / 0.9041 & 33.81 / 0.9427 & -  /  - \\
     & Bicubic \& HAT & $20.9 \mathrm{M}$ & 38.91 / 0.9646 & 35.29 / 0.9293 & 32.74 / 0.9066 & 35.09 / 0.9505 & -  /  - \\
   
    \cline{2-8} & TAD \& TAU & - & 38.46 /  - & 35.52 /  - & 36.68 /  - & 35.03 /  - & 39.01 /  - \\
     & CAR \& EDSR & $52.8 \mathrm{M}$ & 38.94 / 0.9658 & 35.61 / 0.9404 & 33.83 / 0.9262 & 35.24 / 0.9572 & 38.26 / 0.9599 \\
     & IRN & $1.66 \mathrm{M}$ & 43.99 / 0.9871 & 40.79 / 0.9778 & \textcolor{blue}{41.32 / 0.9876} & 39.92 / 0.9865 & 44.32 / 0.9908 \\
     & \textbf{IRRM-S (ours)} & $1.06 \mathrm{M}$ & 44.30 / 0.9895 & 40.95 / 0.9821 & 40.55 / 0.9859 & 40.05 / 0.9885 & 43.98 / 0.9907 \\
     & \textbf{IRRM-M (ours)} & $1.59 \mathrm{M}$ & \textcolor{blue}{44.85 / 0.9904} & \textcolor{blue}{41.39 / 0.9837} & 40.95 / 0.9872 & \textcolor{blue}{40.50 / 0.9896} & \textcolor{blue}{44.44 / 0.9917} \\
     & \textbf{IRRM-L (ours)} & $2.12 \mathrm{M}$ & \textcolor{red}{46.41 / 0.9921} & \textcolor{red}{42.51 / 0.9860} & \textcolor{red}{43.28 / 0.9924} & \textcolor{red}{41.82 / 0.9922} & \textcolor{red}{46.35 / 0.9945} \\
    \hline
    \end{tabular}
    \caption{Quantitative comparison results (PSNR / SSIM) of different rescaling methods on datasets: Set5, Set14, BSD100, Urban100, and DIV2K. -S, -M, and -L are used to denote small, medium, and large parameters of the model, respectively. Best and second best results are in red and blue colors. We report the mean results of 5 draws for IRN, HCFlow and our IRRM. Differences of PSNR and SSIM from different z samples are less than 0.03 and 0.005, respectively.}
    \label{tabs:SOTA}
\end{table*}

        \noindent{\bf Invertible Residual Block.} To be specific, wavelet features $u_1^k$ and $u_2^k$ are fed into stacked IRBs to the obtains LR $y$ and latent  variable $z$. We adopt the general coupling layer for invertible architecture~\cite{i-Revnet,InvertibleResidualNetworks}. The output of each IRB can be defined as:
        \begin{equation}
            \begin{aligned}
                u_1^{k+1} &= u_1^k \cdot  (\text{EB}(u_2^k))+\text{EB}(u_2^k) \\
                u_2^{k+1} &= u_2^k \cdot  (\text{EB}(u_1^{k+1}))+\text{EB}(u_1^{k+1}) \\
                u_2^k &= (u_2^{k+1}-\text{EB}(u_1^{k+1})) /  (\text{EB}(u_1^{k+1})) \\
                u_1^k &= (u_1^{k+1}-\text{EB}(u_2^k)) /  (\text{EB}(u_2^k))
            \end{aligned}
        \end{equation}
        where $u_1^{k+1}$ and $u_2^{k+1}$ are the output of current IRB and the input of next IRB. Note that EB can be arbitrary convolutional unit to enhance the non-linear capability of the model. In our IRRM, we employ two residual connected convolutional blocks and one plain convolutional block~\cite{ResNet,IDN} as the enhanced block, as shown in Fig.~\ref{figs:EB}. RB in EB enhances the non-linearity and the residual connections make the model training stable, solving the problem of unstable training of plain convolutional layers.

    \subsection{Loss Functions}
        Image rescaling aims to reconstruct exactly the HR image while generating visually pleasing LR image. Following IRN~\cite{IRN}, our IRRM is trained by minimizing the following loss:
        \begin{equation}
            \begin{aligned}
                \mathcal{L}= & \lambda_1 \mathcal{L}_{back}(x, x_{back})+\lambda_2 \mathcal{L}_{forw}(y, y_{forw}) \\
                & +\lambda_3 \mathcal{L}_{latent}(z)
            \end{aligned}
        \end{equation}
        where $x$ is ground-truth HR image, $y$ is ground-truth LR image, and $x_{back}$ is reconstructed HR image from forward LR image $y_{forw}$ and sampling latent variable $z$.
        
        $\mathcal{L}_{back}$ is the $l_1$ pixel loss defined as:
        \begin{equation}
            \mathcal{L}_{back}(x_{back}, x)=\frac{1}{M} \sum_{i=1}^M\|f_\theta^{-1}(y, z)-x\|_1
        \end{equation}
        where $M$ is the number of pixels and $x_{back} = f_\theta^{-1}(y, z)$. 
        
        Likewise, $\mathcal{L}_{forw}$ is the $l_2$ pixel loss defined as:
        \begin{equation}
            \mathcal{L}_{forw}(y_{forw}, y)=\frac{1}{N} \sum_{i=1}^N\|f_\theta^y(x)-y\|_2
        \end{equation}
        where $N$ is the number of pixels and $y_{forw} = f_\theta^{y}(x)$ is the ground-truth LR image.

        The last term $\mathcal{L}_{latent}(z)$ is the $l_2$ regularization on the latent variable $z$ to ensure that $z$ follows a Gaussian distribution. We jointly optimise the invertible architecture $f$ by utilizing both forward and backward losses.

        \begin{figure*}[ht]
            \centering
            \includegraphics[width=0.9\linewidth]{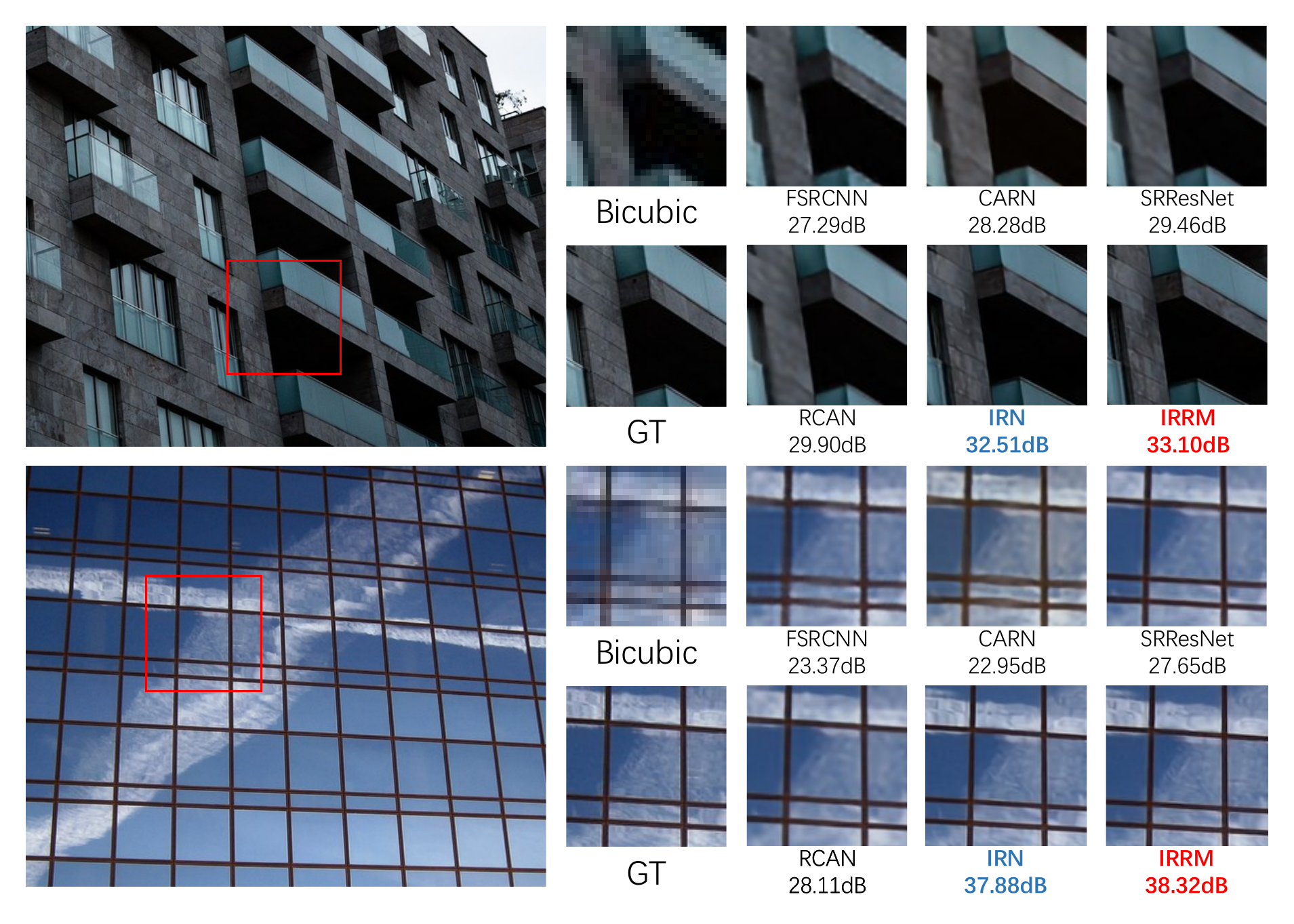}
            \caption{Visual results of upscaling the $4\times$ downscaled images. The right images are $128 \times 128$ which is a patch of the left images. IRRM recovers rich textures and realistic details, leading to better recovery performance. IRRM achieves better performance with an increased PSNR of 4 dB over RCAN and 0.6 dB over IRN}
            \label{figs:show}
        \end{figure*}
        
\section{Experiments}
    \subsection{Settings}
        \noindent{\bf Datasets.} 
        The DIV2K dataset~\cite{DIV2K} is adopted to train our IRRM. Firstly, we prepare the HR images by cropping the original images with steps 240 to $480 \times 480$. These HR images are down-sampled with scaling factors 0.25 and 0.5 to obtain the LR images. Besides, we evaluate our models with PSNR and SSIM  metrics (Y channel) on commonly used benchmarks: Set5~\cite{Set5}, Set14~\cite{Set14}, B100~\cite{B100}, Urban100~\cite{Urban100} and DIV2K valid~\cite{DIV2K} datasets. For a fair comparison with baselines, all images are with size $1280 \times 720$ to calculate FLOPs and we take the mean within a test set.

        \noindent{\bf Training Details.} Our IRRM is composed of one or two Residual Downscaling Modules (RDMs), containing eight Invertible Residual Blocks (IRBs). These models are trained using the ADAM \cite{ADAM} optimizer with $\beta_1 = 0.9$ and $\beta_2 = 0.999$. The batch size is set to 16 for per GPU, and the initial learning rate is fixed at $2 \times 10^{-4}$, which decays by half every 10k iterations. PyTorch is used as the implementation framework. $\lambda_1$, $\lambda_2$ and $\lambda_3$ are set to 8, 8 and 1, respectively. Finally, we augment training images by flipping and rotating.
        
        \begin{figure*}[h]
            \centering
            \includegraphics[width=\linewidth]{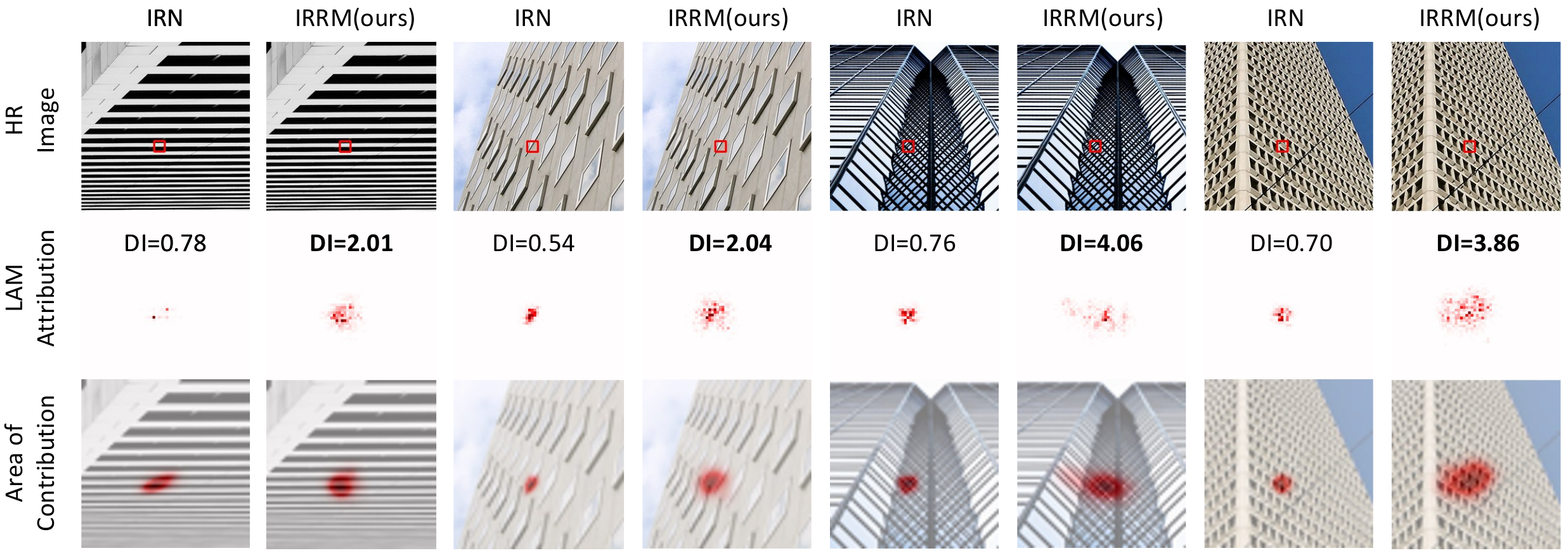}
            \caption{Visual results of Local Attribution Maps (LAM) for IRN and IRRM (ours). The LAM, calculated by diffusion index (DI), indicates the contribution of each pixel in the input image w.r.t. the patch marked with a red box in the HR image. A higher DI, illustrated in the second line, means a wider area of attention in the third lines.}
            \label{figs:LAM}
        \end{figure*}
        \begin{figure*}[t]
            \centering
            \includegraphics[width=0.9\linewidth]{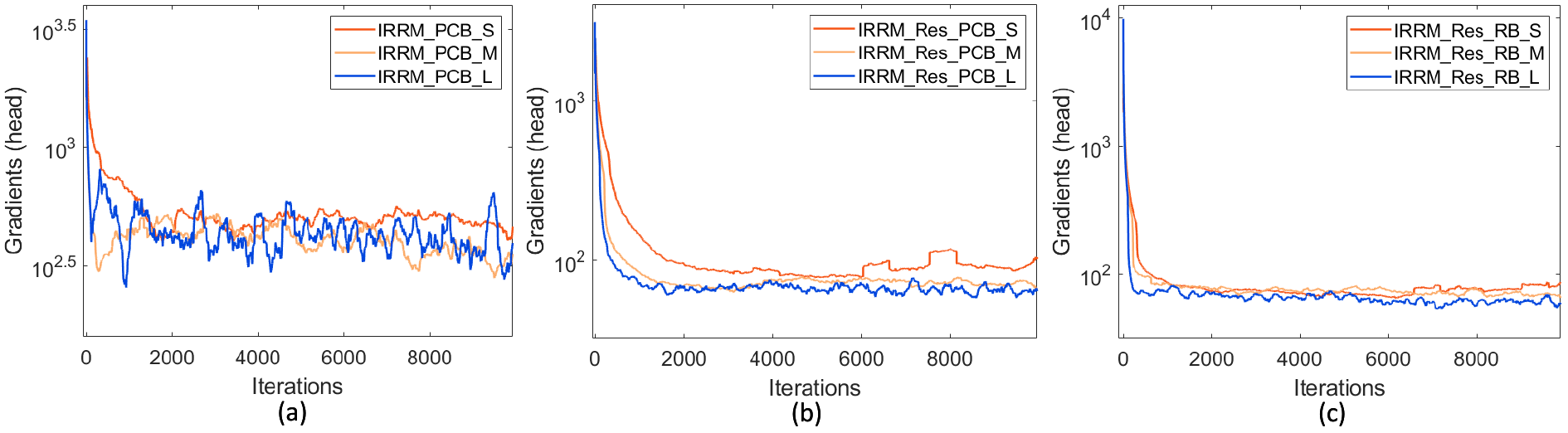}
            \caption{Exploring model extensibility. As the model gets larger, IRRM without residual connections is extremely unstable in training. IRRM with residual connections and RB is stable and converges easily. PCB : plain convolution block. Res : residual connections in the Residual Downscaling Module. RB : residual block in the Invertible Residual Block. -S, -M, -L : denoted the model sizes.}
            \label{figs:gradient_extend}
        \end{figure*}
        
    \subsection{Comparison with State-of-the-art Method}
        \noindent{\bf Performance Comparison.} We compare our IRRM with State-of-the-arts (like HAT~\cite{HAT}, IRN~\cite{IRN}, HCFlow~\cite{HCFlow} and so on) on commonly used datasets (Set5~\cite{Set5}, Set14~\cite{Set14}, B100~\cite{B100}, Urban100~\cite{Urban100} and DIV2K valid~\cite{DIV2K}) evaluated by Peak Signal-to-Noise Ratio (PSNR) and Structural Similarity Index Measure (SSIM).

        As shown in Table~\ref{tabs:SOTA}, the quantitative results of different rescaling methods are summarized. IRRM significantly outperforms previous state-of-the-art methods regarding PSNR and SSIM in five benchmark datasets. Previous SR methods have limited performance in rescaling tasks because of fixed downscaling process. IRN~\cite{IRN}, HCFlow~\cite{HCFlow} and our IRRM optimize the upscaling and downscaling models by joint optimization based on the invertible architecture, further boosting the the PSNR metric about 5 dB on different benchmark datasets. Specifically, HCFlow achieves better performance with an increased PSNR of 0.2 dB over IRN in the $\times 4$ rescaling. Further, compared with IRN, our proposed IRRM improves \textbf{0.6 dB} PSNR in the $\times 4$ rescaling and \textbf{2 dB} PSNR in the $\times 2$ rescaling while only using half of the parameters. IRRM-M achieves performance comparable to IRN with only \textbf{1/4} parameters, and IRRM-S achieves performance well beyond that of previous SR methods with less than \textbf{1M} parameters. These experimental results demonstrate that our model is efficient.

        Besides, we visualize the rescaling resulting images as shown in Fig.~\ref{figs:show}, including two common scenes. HR images restored by IRRM achieve better visual quality and rich textures than those of previous state-of-the-art methods. This advanced capability is attributed to residual connections in RDM, which enables the model to concentrate on learning high-frequency information, and the incorporation of Invertible Residual Block (IRB), which strengthens non-linear capacity of the model.
        
        \noindent{\bf Interpretation with Local Attribution Maps (LAM).} We analyze the diffusion index (DI) of previous methods and IRRM to better understand why IRRM works. DI is applied to measure local attribution maps (LAM~\cite{LAM}) from the input image to output image of models. When considering the same local patch, if LAM contains a greater range of pixels or encompasses a larger area, it can be inferred that the models have effectively utilized information from a larger number of pixels. As shown in Fig.~\ref{figs:LAM}, our IRRM utilizes a wider range of pixels for better rescaling for models. Specifically, the DI of IRRM is four times that of IRN due to the fact that we have introduced residual connections in IRRM, which makes the receptive field of the model larger.       

        \begin{table}[h]
            \centering
            \begin{tabular}{c|c|c|c|c|c}
            \hline RDM & IRB & Set5 & Set10 & B100 & Urban100 \\
            \hline \XSolidBrush & PCB & 35.23 & 31.55 & 30.77 & 29.63 \\
             \XSolidBrush & RB & \textbf{36.56}& \textbf{31.90} & \textbf{31.03} & \textbf{30.25} \\
             
            \hline \CheckmarkBold & PCB & 36.80 & 33.27 & 32.22 & 31.68 \\
             \CheckmarkBold & RB  & \textbf{36.85} & \textbf{33.37} & \textbf{32.30} & \textbf{31.94} \\
             \CheckmarkBold & RB+ & 36.46 & 32.81 & 31.76 & 31.23 \\
             
            \hline \XSolidBrush   & 4 RBs & 35.07 & 31.41 & 30.57 & 29.51 \\
             \CheckmarkBold & 4 RBs & 36.30 & 32.84 & 31.93 & 31.36 \\
             \XSolidBrush   & 8 RBs & 35.56 & 31.90 & 31.03 & 30.25 \\
             \CheckmarkBold & 8 RBs & 36.85 & 33.37 & 32.30 & \textbf{31.94} \\
             \XSolidBrush   & 12 RBs & 35.48 & 31.90 & 30.87 & 29.93 \\
             \CheckmarkBold & 12 RBs & \textbf{36.89} & \textbf{33.41} & \textbf{32.32} & \textbf{31.94} \\
            \hline
            \end{tabular}
            \caption{Ablation study of the effect of proposed Residual Downscaling Module (RDM) and Invertible Residual Block (IRB). The first column indicates whether residual connections are used in the RDM. The second column indicates the convolutional blocks in the IRB. N RBs denotes the IRB uses n RBs.}
            \label{tab:RDM_IRB}
        \end{table}
    \subsection{Ablation Study}
        In the ablation study, we replace the modules proposed in the IRRM to evaluate the effect of RDM and IRB. Model extensibility and effects of loss functions are also evaluated.
        
        \noindent{\bf Residual Downscaling Module.} We discard residual connections of RDM to investigate the contribution of them. As shown in Fig.~\ref{figs:gradient_extend}, IRRM is more extensible with residual connections and RB. As the model gets deeper, the absence of residual connections can lead to instability of the model gradient, with problems of gradient vanishing and gradient explosion. Furthermore, IRRM with residual connections has better extensibility while the performance degrades as the network gets larger (8 RBs $\rightarrow$ 12 RBs) without residual connections, as shown below in Table~\ref{tab:RDM_IRB}.

        \noindent{\bf Invertible Residual Block.} To investigate the effects of Residual Block (EB) in the IRB, we replace it with a plain convolutional layer. The detailed structure is shown in Fig.~\ref{figs:EB}. As shown above in Table~\ref{tab:RDM_IRB}, RB performs better than PCB, due to residual connections. In addition, RB+ achieves comparable performance through half of the parameters. 

        As illustrated in Table~\ref{tab:extensibility}, the performance of IRRM with residual connections and RB improves as the model size increases (IRRM-S (RB, Res) $\rightarrow$ IRRM-M (RB, Res) $\rightarrow$ IRRM-L (RB, Res)). However, in the case of IRRM without residual connections, the performance does not improve as the model size increases (IRRM-S (PCB) $\rightarrow$ IRRM-M (PCB) $\rightarrow$ IRRM-L (PCB)). Notably, IRRM-M (PCB) achieves superior performance in this scenario. This shows the excellent extensibility of our proposed IRRM.
        
        \begin{table}[t]
            \small
            \centering
            \begin{tabular}{c|c|c|c|c}
                \hline Models & Set5 & Set10 & B100 & Urban100 \\
                \hline IRRM-S (PCB) &  34.86 & 31.21 & 30.43 & 29.09 \\
                 IRRM-M (PCB) &  \textbf{35.23} & \textbf{31.55} & \textbf{30.77} & 29.63 \\
                 IRRM-L (PCB) &  34.68 & 31.53 & 30.73 & \textbf{29.66} \\
                 
                \hline IRRM-S (PCB, Res) &  36.09 & 32.42 & 31.68 & 31.14 \\
                 IRRM-M (PCB, Res) &  \textbf{36.80} & 33.27 & \textbf{32.22} & 31.68 \\
                 IRRM-L (PCB, Res) &  36.72 & \textbf{33.29} & 32.19 & \textbf{31.71} \\
                 
                \hline IRRM-S (RB, Res) &  36.30 & 32.87 & 31.88 & 31.30 \\
                 IRRM-M (RB, Res) &  36.85 & 33.37 & 32.30 & 31.94 \\
                 IRRM-L (RB, Res) &  \textbf{36.89} & \textbf{33.41} & \textbf{32.32} & \textbf{31.94} \\
                \hline
            \end{tabular}
            \caption{Exploring different model extensibility. IRRM (PCB): IRRM with Plain convolutional block. IRRM (PCB, Res): IRRM with plain convolutional blocks and residual connections. IRRM (RB, Res): IRRM with residual blocks and residual connections.}
            \label{tab:extensibility}
        \end{table}

    \begin{figure}[t]
        \centering
        \includegraphics[width=\linewidth]{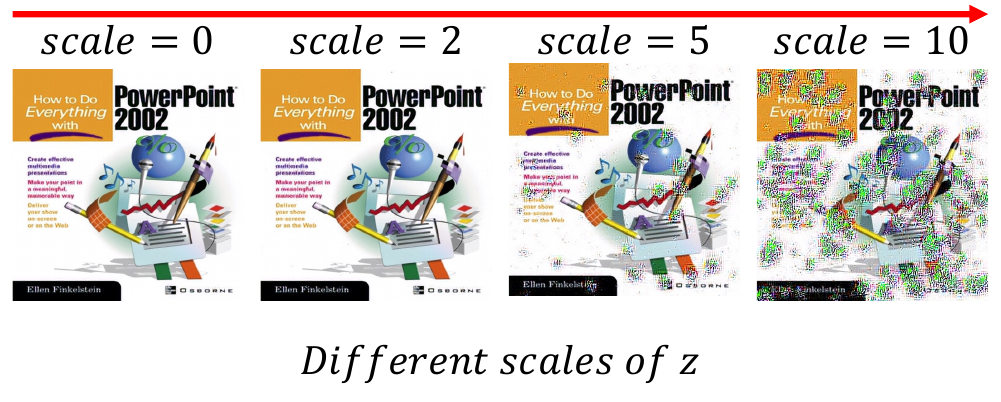}
        \caption{Visual results of employing IRRM to HR images using different scales of $z$ samples. A Gaussian distribution is used in training $z$, and scaled samples drawn from this distribution are used to illustrate upscaling results.}
        \label{figs:Ztest}
    \end{figure}
    
    \subsection{Visualisation on the Influence of $z$}
        We also analyse the effect of different scales of z on the reconstructed HR image, as shown in Fig.~\ref{figs:Ztest}. It is observed that our proposed IRRM is insensitive to the Gaussian distribution z and generates unrealistic high-frequency details only when z is very large. $z$ represents the high-frequency components in an image and is typically sparse. However, changes in $z$ have a significant impact on the image details. Therefore, it is essential for the model to exhibit a high tolerance towards $z$ in order to preserve image details effectively.

    \subsection{Real-world Datasets}
        Real-world dataset is also applicable for our method, since the inputs of our model are high-resolution images, whether they are real-world datasets or synthetic datasets. The results are shown in Table~\ref{tab1}.
        \begin{table}[ht]
            \small
            \centering
            \begin{tabular}{c|c|c|c|c}
                \hline
                \multirow{2}{*}{ Method } & \multicolumn{2}{|c|}{ RealSR-cano } & \multicolumn{2}{c}{ RealSR-Nikon } \\
                \cline { 2 - 5 } & PSNR $\uparrow$ & LPIPS $\downarrow$ & PSNR $\uparrow$ & LPIPS $\downarrow$ \\
                \hline 
                ESRGAN & 27.67 & 0.412 & 27.46 & 0.425 \\
                SwinIR & 26.64 & 0.357 & 25.76 & 0.364 \\
                HAT & 26.68 & 0.342 & 25.85 & 0.358 \\
                \textbf{Ours} & \textbf{29.12} & \textbf{0.327} & \textbf{28.87} & \textbf{0.336} \\
                \hline
            \end{tabular}
            \caption{Comparison results with existing state-of-the-art SR methods on real-world datasets.}
            \label{tab1}
        \end{table}

        With the recent rapid development in multimodality~\cite{bai2023badclip,li2024fmm,gao2023backdoor,wang2023contrastive}, we will also explore the help of textual cues in the future to better exploit the potential of the Rescaling task.
\section{Conclusion}
    In this paper, we develop an efficient and light-weight framework IRRM for image rescaling. IRRM learns a bijection between HR image and LR image as well as specific distribution $z$ so that it is information lossless and invertible. Specifically, IRRM introduces residual connections in the RDM, equivalent to a second-order wavelet transform. Direct modeling of high frequency information reduces the training difficulty and allows the training gradient to trend towards brown noise rather than white noise. Besides, we propose IRB to enhance non-linear representation and reduce model degradation, while the block is lightweight and lossless. This addition significantly improves the extensibility of the model.

\section{Acknowledgements}
    This work is supported in part by the National Natural Science Foundation of China, under Grant (62302309, 62171248),  Shenzhen Science and Technology Program (JCYJ20220818101014030, JCYJ20220818101012025), and the PCNL KEY project (PCL2023AS6-1), and Tencent ``Rhinoceros Birds" - Scientific Research Foundation for Young Teachers of Shenzhen University.

\newpage
\bibliographystyle{named}
\bibliography{main}

\end{document}